\documentclass[10pt,twocolumn,letterpaper]{article}

\usepackage{iccv}
\usepackage{times}
\usepackage{epsfig}
\usepackage{graphicx}
\usepackage{amsmath}
\usepackage{amssymb}
\usepackage{multirow}
\usepackage{pifont}

\usepackage[pagebackref=true,breaklinks=true,letterpaper=true,colorlinks,bookmarks=false]{hyperref}

\iccvfinalcopy 

\newcommand{\cmark}{\ding{51}}%
\newcommand{\xmark}{\ding{55}}%
\newcommand{\tabincell}[2]{\begin{tabular}{@{}#1@{}}#2\end{tabular}}

\def\iccvPaperID{5842} 

\ificcvfinal\fi

\begin{document}

\title{MonoIndoor: Towards Good Practice of Self-Supervised \\ Monocular Depth Estimation for Indoor Environments}

\author{Pan Ji\thanks{Joint first authorship. P. Ji is the corresponding author (peterji530@gmail.com). R. Li's contribution was made during an internship with OPPO US Research Center.}\;\textsuperscript{1}\;, Runze Li\footnotemark[1]\;\textsuperscript{1,2}\;,  Bir Bhanu\textsuperscript{2}, Yi Xu\textsuperscript{1}\\
\textsuperscript{1}OPPO US Research Center, InnoPeak Technology, Inc. \\ \textsuperscript{2}University of California Riverside\\
}

\maketitle
\ificcvfinal\thispagestyle{empty}\fi

\begin{abstract}
    Self-supervised depth estimation for indoor environments is more challenging than its outdoor counterpart in at least the following two aspects: (i) the depth range of indoor sequences varies a lot across different frames, making it difficult for the depth network to induce consistent depth cues, whereas the maximum distance in outdoor scenes mostly stays the same as the camera usually sees the sky; (ii) the indoor sequences contain much more rotational motions, which cause difficulties for the pose network, while the motions of outdoor sequences are pre-dominantly translational, especially for driving datasets such as KITTI. In this paper, special considerations are given to those challenges and a set of good practices are consolidated for improving the performance of self-supervised monocular depth estimation in indoor environments. The proposed method mainly consists of two novel modules, \ie, a depth factorization module and a residual pose estimation module, each of which is designed to respectively tackle the aforementioned challenges. The effectiveness of each module is shown through a carefully conducted ablation study and the demonstration of the state-of-the-art performance on three indoor datasets, \ie, EuRoC, NYUv2 and 7-Scenes.
\end{abstract}

\section{Introduction}
\label{sec:intro}
Depth estimation plays an essential role in a variety of 3D perceptual tasks, such as autonomous driving, virtual reality (VR), and augmented reality (AR). In this paper, we tackle the problem of estimating the depth map from a single image in a self-supervised manner. Compared to the supervised methods~\cite{Eigen_2015_ICCV,Fu_2018_CVPR}, self-supervision~\cite{garg2016unsupervised, zhou2017unsupervised, godard2019digging} frees us from having to capture the ground-truth depth using depth sensors (\eg, LiDAR) and therefore, it is more attractive in scenarios where obtaining the ground-truth is not possible.

Recently, self-supervised methods~\cite{godard2019digging} have achieved significant success, producing depth prediction that is comparable to that produced by the supervised methods~\cite{guo2018learning,Fu_2018_CVPR}. For example, on the KITTI dataset~\cite{Geiger2012CVPR}, Monodepth2~\cite{godard2019digging} achieves an absolute relative depth error (AbsRel) of 10.6\%, which is not far from the AbsRel of 7.2\% by supervised DORN~\cite{Fu_2018_CVPR}. However, most of these self-supervised depth prediction methods~\cite{garg2016unsupervised,zhou2017unsupervised,godard2019digging} are only evaluated on outdoor datasets such as KITTI, leaving their performance opaque 
for indoor environments. A few methods~\cite{zhou2019moving,zhao2020towards} 
have considered indoor self-supervised depth prediction, but their performance still trail far behind the one on the outdoor datasets by methods such as~\cite{garg2016unsupervised,zhou2017unsupervised,godard2019digging} 
or the supervised counterparts~\cite{Fu_2018_CVPR,Yin_2019_ICCV} on indoor datasets. For instance, on the indoor NYUv2 dataset~\cite{Silberman:ECCV12}, the method by Zhao~\etal~\cite{zhao2020towards} reaches an AbsRel of 18.9\%, which is much higher than what Monodepth2 can achieve on KITTI. 

In view of the performance discrepancies between the indoor and outdoor scenes, we examine what makes indoor depth prediction more challenging than the outdoor case. Our first conjecture is that this is partly due to the fact that the scene depth range of indoor sequences varies a lot more than in the outdoor. This results in more difficulties for the depth network in inducing consistent depth cues across images. Our second observation is that the pose network, which is commonly used in self-supervised methods~\cite{zhou2017unsupervised,godard2019digging}, tends to have large errors in rotation prediction. A similar finding in ~\cite{zou2020learning} shows that predicted poses have much higher rotational errors (\eg, 10 times larger) than geometric SLAM~\cite{mur2017orb} even after using a recurrent pose network. This problem is not prominent on KITTI because the motions therein are mostly translational. However, since indoor datasets are often captured by hand-held cameras~\cite{Silberman:ECCV12} or MAVs~\cite{schonberger2016structure} which inevitably undergo frequent rotations, the inaccurate rotation prediction becomes detrimental to the self-supervised training of a depth model for indoor environments.

Given the above considerations, we propose {\bf MonoIndoor}, a monocular self-supervised depth estimation method tailored for indoor environments. Our MonoIndoor 
consists of two novel modules: a \textit{depth factorization} module and a \textit{residual pose estimation} module. In the depth factorization module, we factorize the depth map into a global depth scale (for the current image) and a relative depth map. The depth scale factor is separately predicted by an extra branch in the depth network. In such a way, the depth network has more model plasticity to adapt to the depth scale changes during training. In the residual pose estimation module, we mitigate the issue of inaccurate rotation prediction by performing residual pose estimation in addition to an initial large pose prediction. Such a residual approach leads to more accurate computation of the photometric loss~\cite{godard2019digging}, which in turn leads to better model training for the depth network.

In summary, our contributions are:
\begin{itemize}
    \setlength\itemsep{0cm}
    \item A novel depth factorization module that helps the depth network adapt to the rapid scale changes;
    \item A novel residual pose estimation module that mitigates the inaccurate rotation prediction issue in the pose network and in turn improves depth prediction;
    \item State-of-the-art performance of self-supervised depth prediction on three publicly available indoor datasets, \ie, EuRoC~\cite{schonberger2016structure}, NYUv2~\cite{Silberman:ECCV12}, and 7-Scenes~\cite{Shotton_2013_CVPR}.
\end{itemize}

\section{Related Work}
In this section, we review both supervised and self-supervised methods for monocular depth estimation.

\subsection{Supervised Monocular Depth Estimation}
Early depth estimation methods are mostly supervised. Saxena~\etal~\cite{saxena2008make3d} regress the depth from a single image with superpixel features and a Markov Random Field (MRF). Eigen~\etal~\cite{eigen2014depth} propose the first deep-learning based method for monocular depth estimation using a multi-scale convolutional neural network (CNN). Later methods improve the performance of depth prediction either by better network architecture~\cite{laina2016deeper} or via more sophisticated training losses~\cite{Li_2017_ICCV,Fu_2018_CVPR,Yin_2019_ICCV}. A few methods~\cite{ummenhofer2017demon,teed2018deepv2d} rely on two networks, one for depth prediction and the other for motion, to mimic geometric Structure-from-Motion (SfM) or Simultaneous Localization and Mapping (SLAM) in a supervised framework. Training these methods needs ground-truth depth data, which is often expensive to capture. Some other methods then resort to generating pseudo ground-truth depth labels with traditional 3D reconstruction methods~\cite{li2018megadepth,li2019learning}, such as SfM~\cite{schonberger2016structure} and SLAM~\cite{mur2017orb}, or 3D movies~\cite{ranftl2019towards}. Such methods have better capacity of generalization across different datasets, but can not necessarily achieve the best performance for the dataset at hand.

\subsection{Self-Supervised Monocular Depth Estimation}
Self-supervised depth estimation has attracted a lot of attention recently as it does not require training with the ground truth. Along this line, Garg~\etal~\cite{garg2016unsupervised} propose the first self-supervised method to use color consistency loss between stereo images to train a monocular depth model. Zhou~\etal~\cite{zhou2017unsupervised} employ two networks (\ie, one depth network and one pose network) to construct the photometric loss across temporal frames. Many follow-up methods then try to improve the self-supervision by new loss terms. Godard~\etal~\cite{godard2017unsupervised} incorporate a left-right depth consistency loss for the stereo training. Bian~\etal~\cite{bian2019unsupervised} put forth a temporal depth consistency loss to encourage neighboring frames to have consistent depth predictions. Wang~\etal~\cite{wang2018learning} observe the diminishing issue of the depth model during training and come up with a simple normalization method to counter this effect. Yin~\etal~\cite{yin2018geonet} and Zou~\etal~\cite{zou2018dfnet} use three networks (\ie, one depth network, one pose network, and one extra flow network) to enforce cross-task consistency between optical flow and dense depth. Wang~\etal~\cite{wang2019recurrent} and Zou~\etal~\cite{zou2020learning} leverage recurrent neural networks, such as LSTMs, to model long-term dependency in the pose network and/or the depth network. Tiwari~\etal~\cite{tiwari2020pseudo} form a self-improving loop with monocular SLAM and a self-supervised depth model~\cite{godard2019digging} to improve the performance of each one. Notably, Monodepth2~\cite{godard2019digging} significantly improves the performance over previous methods via a set of techniques: a per-pixel minimum photometric loss to handle occlusions, an auto-masking method to mask out static pixels, and a multi-scale depth estimation strategy to mitigate the texture-copying issue in depth. Due to its good performance, we implement our self-supervised depth estimation framework based on Monodepth2, but make important changes to both the depth and the pose networks.

Most of the aforementioned methods are only evaluated on outdoor datasets such as KITTI. A few other recent methods~\cite{zhou2019moving,zhao2020towards,bian2020unsupervised} focus on indoor self-supervised depth estimation. Zhou~\etal~\cite{zhou2019moving} propose an optical-flow based training paradigm and handle large rotational motions by a pre-processing step that removes all the image pairs with ``pure rotation''. Zhao~\etal~\cite{zhao2020towards} adopt a geometry-augmented strategy that solves for the depth via two-view triangulation and then uses the triangulated depth as supervision. Bian~\etal~\cite{bian2020unsupervised} argue that ``the rotation behaves as noise during training'' and thus propose a 
rectification step to remove the rotation between consecutive frames. We have an observation similar to~\cite{zhou2019moving} and~\cite{bian2020unsupervised} that large rotations cause difficulties for the network. However, we take a different strategy. Instead of removing rotations from training data, we progressively estimate them via a novel residual pose module. This in turn improves depth prediction.

\section{Method}

\begin{figure*}[ht]
\begin{center}
\includegraphics[width=1.0\textwidth]{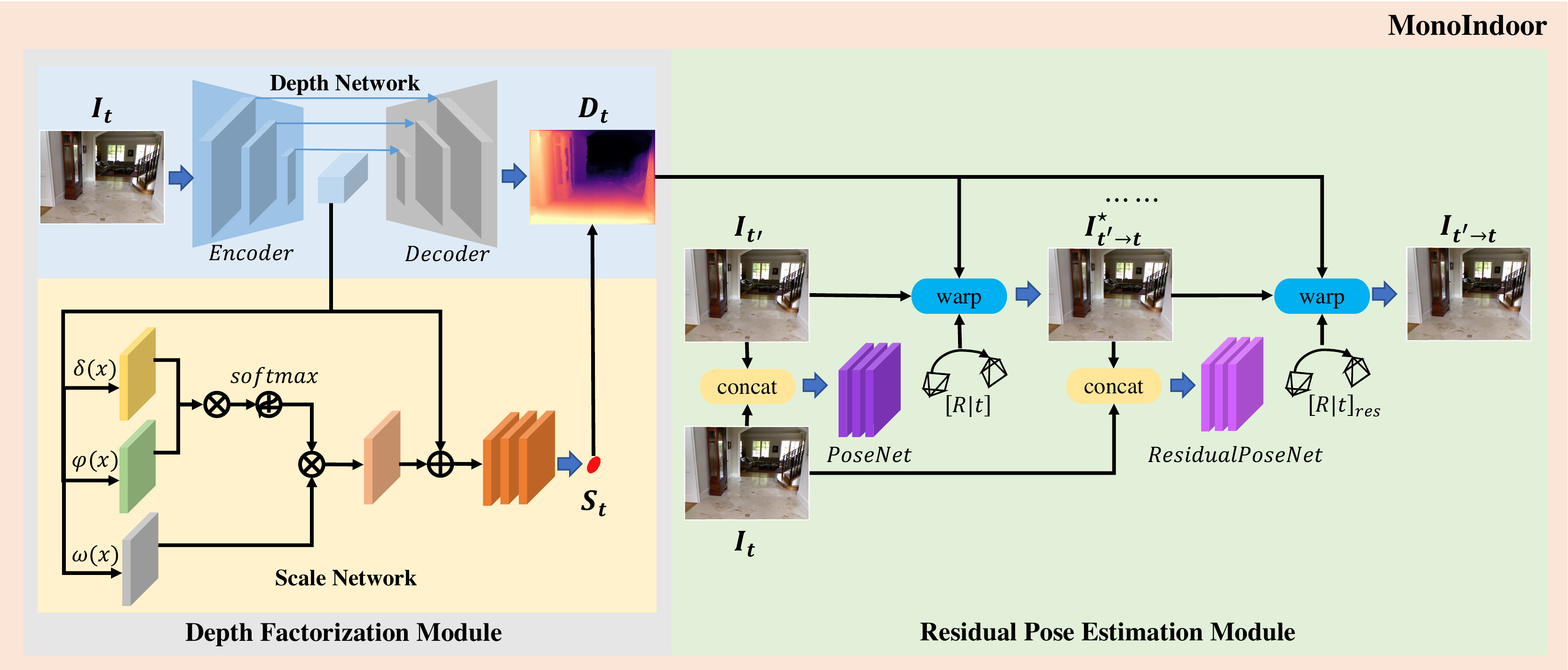}
\end{center}
\caption{Overview of the proposed \textbf{MonoIndoor}. \textbf{Depth Factorization Module}: We use an encoder-decoder based depth network to predict a relative depth map and a non-local scale network to estimate a global scale factor. \textbf{Residual Pose Estimation Module}: We use a pose network to predict an initial camera pose of a pair of frames and residual pose network to iteratively predict residual camera poses based on the predicted initial pose.}
\label{fig:pipeline}
\end{figure*}

In this section, we give detailed descriptions of performing self-supervised depth estimation using \textbf{MonoIndoor}. Specifically, we first introduce the background of the self-supervised depth estimation. Then, we describe the good practices in predicting depth with our MonoIndoor. 

\subsection{Self-Supervised Depth Estimation}
\label{depth_estimation}
Similar to~\cite{zhou2017unsupervised,godard2019digging,zou2020learning}, we also consider the self-supervised depth estimation as a novel view-synthesis problem by training a model to predict the target image from different viewpoints of source images. The image synthesis process is trained and constrained by using the depth map as the bridging variable. Such a system requires both the predicted depth map of the target image and the estimated relative pose between a pair of target and source images. Specifically, given a target image $I_t$ and a source image $I_{t'}$ from another view, the system is jointly trained to predict a dense depth map ${D}_t$ of the target image and a relative camera pose ${T}_{t \rightarrow t'}$ from the target to the source. The photometric reprojection loss can then be constructed as follows:
\begin{equation}
    \mathcal{L}_{A} = \sum_{t'}\rho(I_t, I_{t'\rightarrow t}),
\end{equation}
and 
\begin{equation}
    I_{t'\rightarrow t} = I_{t'}\langle proj({D}_t, {T}_{t\rightarrow t'}, K) \rangle,
\label{eq:warp}
\end{equation}
where $\rho$ denotes the photometric reconstruction error~\cite{zhou2017unsupervised, godard2019digging}. It is a weighted combination of the L1 and Structured SIMilarity (SSIM) loss defined as 
\begin{equation}
    \rho(I_t, I_{t'\rightarrow t}) = \frac{\alpha}{2}\big(1-\texttt{\footnotesize SSIM}(I_t, I_{t'\rightarrow t})\big) + (1-\alpha)\|I_t, I_{t'\rightarrow t}\|_1.
    \label{eq:ssim_l1_loss}
\end{equation}
$I_{t'\rightarrow t}$ is the source image warped to the target coordinate frame based on the depth of the target image. $proj()$ is the transformation function to map image coordinated $p_{t}$ from the target image to its $p_{t'}$ on the source image following
\begin{equation}
    p_{t'} \sim K{T}_{t\rightarrow t'}{D}_t(p_t)K^{-1}p_t,
    \label{eq:transform}
\end{equation}
and $\langle\cdot\rangle$ is the bilinear sampling operator which is locally sub-differentiable. Following~\cite{godard2019digging}, the camera intrinsics $K$ of all images are assumed to be the same, and an edge-ware smoothness term is employed as 
\begin{equation}
    \mathcal{L}_{s} = |\partial_{x}d^{\ast}_{t}|e^{-|\partial_{x}I_t|}+|\partial_{y}d^{\ast}_{t}|e^{-|\partial_{y}I_t|},
\end{equation}
where $d^{\ast}_t = d/\bar{d}_t$ is the mean-normalized inverse depth from~\cite{wang2018learning}. During training, we adopt the auto-masking scheme~\cite{godard2019digging} to handle static pixels.

Similar to~\cite{bian2019unsupervised}, we use an additional depth consistency loss to enforce consistent depth prediction across neighboring frames. We first warp the depth image ${D}_{t'}$ of the source image by Equation~\eqref{eq:warp} to generate ${D}_{t'\rightarrow t}$, which is a corresponding depth map in the coordinate system of the source image. We then transform ${D}_{t'\rightarrow t}$ to the coordinate system of the target image via Equation~\eqref{eq:transform} to produce a synthesized target depth map $\widetilde{D}_{t'\rightarrow t}$. The depth consistency loss can be written as
\begin{equation}
    \mathcal{L}_c = \frac{|{D}_t - \widetilde{D}_{t'\rightarrow t}|} {{D}_t + \widetilde{D}_{t'\rightarrow t}}.
\end{equation}


The overall objective to train the model is
\begin{equation}
    \mathcal{L} = \mathcal{L}_{A} + \tau\mathcal{L}_{s} + \gamma \mathcal{L}_c,
\label{eq:photo_loss}
\end{equation}
where $\tau$ and $\gamma$ are the weights for the edge-aware smoothness loss and the depth consistency loss respectively.

Even though existing monocular self-supervised methods are able to produce competitive depth maps in outdoor environments, these methods still suffer from worse performance in indoor environments, especially compared with fully-supervised methods. As discussed in Section~\ref{sec:intro}, the main challenges in indoor environments come from the fact that the depth range changes a lot and indoor sequences contain regular rotational motions which are difficult to predict. To handle these issues, we propose \textbf{MonoIndoor}, a monocular self-supervised depth estimation framework, as shown in Figure~\ref{fig:pipeline}, to enable improved predicted depth quality in indoor environments.

The system takes as input a single color image and outputs a depth map via our MonoInoor which consists of two core parts: a depth factorization module and a residual pose estimation module. 
We present our main contributions in the following sections.

\subsection{Depth Factorization}
\label{sec:depth_scale_factorization}
We use the Monodepth2~\cite{godard2019digging} as the backbone model for depth prediction. The depth model in Monodepth2 employs an auto-encoder structure with skip connections between the encoder and the decoder. The depth encoder takes as input a color image $I$, and the decoder outputs its depth map. 
Note that the final depth prediction is not directly from the convolutional layers, but after a sigmoid activation function and a linear scaling function as follows,
\begin{equation}
d = 1/(a\sigma + b), 
\label{eq:depth_factor}
\end{equation}
where $\sigma$ is the value after the sigmoid function, $a$ and $b$ are specified to constrain the depth map ${D}$ within a certain depth range. Practically, $a$ and $b$ are respectively pre-defined as a minimum depth value and a maximum depth value which can be obtained in a known environment. For instance, on the KITTI dataset ~\cite{Geiger2012CVPR}, $a$ is chosen as 0.1 and $b$ as 100. The reason for setting $a$ and $b$ as fixed values is that the depth range is consistent across the video sequences when the camera always sees the sky at the far point. However, this setting is not valid for most indoor environments. As scene varies, the depth range varies a lot. For example, the depth range in a bathroom (\eg 0.1m$\sim$3m) can be very different from the one in a lobby (\eg 0.1m$\sim$10m). Pre-setting depth range will act as an inaccurate guidance that is harmful for the model to capture accurate depth scales. This is especially true when there are rapid scale changes, which are commonly observed in indoor scenes. To overcome this problem, we propose a depth factorization module (see Figure~\ref{fig:pipeline}) to learn a disentangled representation in the form of a relative depth map and a global scale factor. 
We employ the depth network of Monodepth2~\cite{godard2019digging} to predict relative depth and propose a self-attention-guided scale regression network to predict the global scale factor for the current view.

\noindent\textbf{Scale Network.} We design the scale network as a new branch which takes as input a color image and outputs its global scale factor. 
Since the global scale factor is closely informed by certain areas (\eg, the far point) in the images, we explore to use a self-attention block~\cite{Wang_nonlocalCVPR2018} so that the network can be guided to pay more attention to a certain area which is informative to induce the depth scale factor of the current view in a scene. 
Given the feature representations $\mathcal{F}$ learnt from the input image, we utilize a self-attention block to take $\mathcal{F}$ as input, forming the query, the key and the value output by
\begin{equation}
  \begin{aligned}
  \psi(\mathcal{F}) &= \mathbf{W}_{\psi}\mathcal{F},
  \\
  \phi(\mathcal{F}) &= \mathbf{W}_{\phi}\mathcal{F},
  \\
  h(\mathcal{F}) &= \mathbf{W}_{h}\mathcal{F},
  \end{aligned}
\end{equation}
where $\mathbf{W}_{\psi}$, $\mathbf{W}_{\phi}$ and $\mathbf{W}_{h}$ are parameters to be learnt. The query and key values are then combined in $\mathcal{G_F}=\texttt{softmax}(\mathcal{F}^{T}\mathbf{W}^{T}_{\psi}\mathbf{W}_{\phi}\mathcal{F})h(\mathcal{F})$ as the learnt self-attentions. Finally, the self-attention $\mathcal{G_F}$ and $\mathcal{F}$ jointly contribute to the output $\mathcal{S_F}$ by using 
\begin{equation}
  \mathcal{S_F} = \mathbf{W}_{\mathcal{S_F}}\mathcal{G_F} + \mathcal{F}.
\end{equation}
Once we obtain the attentive representations as $\mathcal{S_F}$, we apply two residual blocks including two convolutional layers in each, followed by three fully-connected layers with dropout layers in-between, to output the global scale factor $S$ for the current image.

\noindent\textbf{Probabilistic Scale Regression Head.}
To predict a global scale, a high-dimensional feature map has to be mapped into a single positive number. One straightforward way is to let the network directly regress the scale number.
However, we observe unstable training using this approach. To mitigate this issue, inspired by~\cite{chang2018pyramid}, we propose to use a probabilistic scale regression head to estimate this continuous value. Given a maximum bound that the global scale factor is within, the probability of each scale $s$ is calculated from the output of the scale network $\mathcal{\widetilde{S}}$ via the softmax operation $\texttt{softmax}(\cdot)$. The predicted global scale ${S}$ is calculated as the sum of each scale $s$ weighted by its probability as
\begin{equation}
  {S} = \sum^{D_{max}}_{s=0}s\times \texttt{softmax}(\mathcal{\widetilde{S}}).
\end{equation}
By doing so, the regression problem is smoothly resolved by a probabilistic classification-based strategy (see Section~\ref{sec:euroc_ablation} for more ablation results). 

\subsection{Residual Pose Estimation}
As mentioned in Section~\ref{depth_estimation}, self-supervised depth estimation is built upon the novel view synthesis, which requires both accurate depth maps and camera poses. Estimating accurate relative poses is key for the photometric reprojection loss because inaccurate poses might lead to wrong correspondences between the target and source pixels, causing problems in predicting the depth. 
Existing methods mostly employ a standalone PoseNet to estimate the 6 Degrees-of-Freedom (DoF) pose between two images. In outdoor environments (\eg, driving scenes like KITTI), the relative camera poses are fairly simple because the cars are mostly moving forward with large translations but minor rotations. This means that pose estimation is normally less challenging. In contrast, in indoor environments, the sequences are typically recorded with hand-held devices (\eg, Kinect), so there are more complicated ego-motions involved as well as much larger rotational motions. It is thus more difficult for the pose network to learn accurate camera poses. 

\begin{figure}[!t]
\begin{center}
\includegraphics[width=\linewidth]{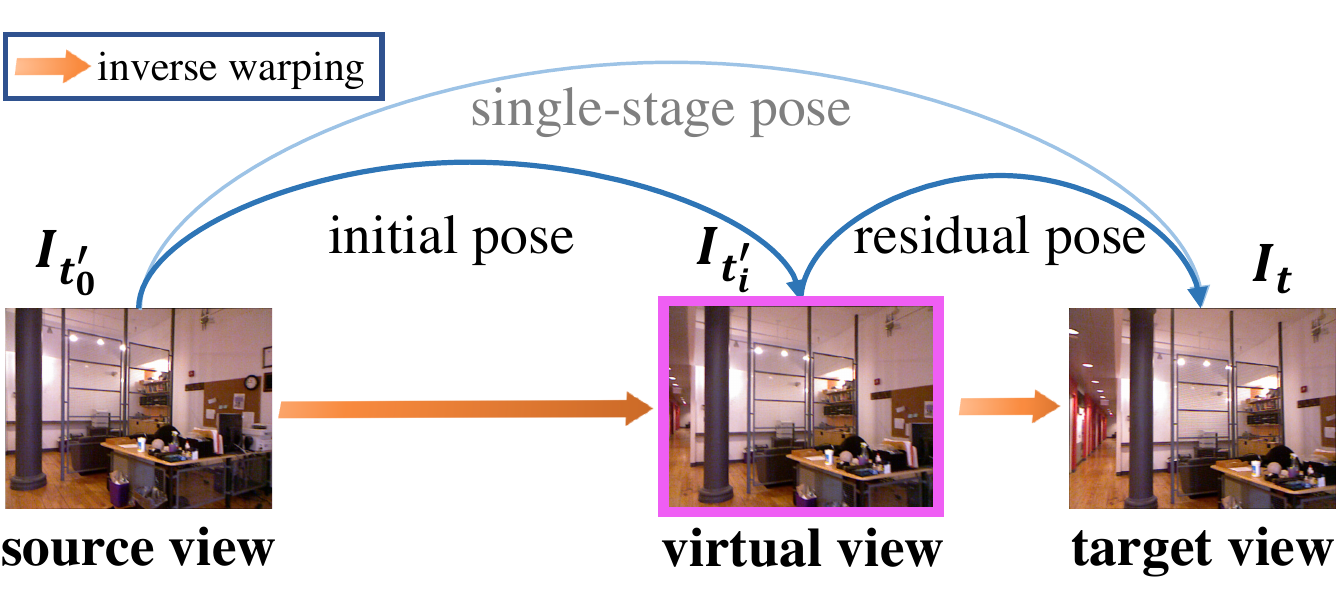}
\end{center}
\caption{Residual Pose Estimation. Here we give an illustrative example of how a single-stage pose can be decomposed into an {\it initial pose} and a {\it residual pose} by virtual view synthesis.}
\label{fig:residual_pose_pipeline}
\end{figure}

Unlike existing methods~\cite{zhou2019moving,bian2020unsupervised} that concentrate on ``removing" or ``reducing" rotational components during data preprocessing, 
we instead propose a residual pose estimation module to learn the relative camera pose between the target and source images in an iterative manner (see Figure~\ref{fig:residual_pose_pipeline}). 
In the first stage, the pose network takes a target image $I_t$ and a source image $I_{t'_0}$ as input and predicts an initial camera pose ${T}_{t'_0\rightarrow t}$, where the subscript 0 in $t'_{0}$ indicates that no transformation is applied yet. 
We then follow Equation~\eqref{eq:warp} to bilinearly sample from the source image, reconstructing a virtual view $I_{t'_{0}\rightarrow t}$ which is expected to be the same as the target image $I_t$ if the correspondences match accurately. However, it will not be the case due to inaccurate pose prediction. Note here the transformation is defined as
\begin{equation}
    I_{t'_0\rightarrow t} = I_{t'}\langle proj({D}_t, {T}^{-1}_{t'_{0}\rightarrow t}, K) \rangle.
\label{eq:res_warp}
\end{equation}
Next, we utilize a residual pose network (see \textbf{\textit{ResidualPoseNet}} in Figure~\ref{fig:pipeline}) which takes the target image and the synthesized view $I_{t'_{0}\rightarrow t}$ as input and outputs a residual camera pose ${T}^{res}_{(t'_{0}\rightarrow  t)\rightarrow t}$, representing the camera pose 
of the synthesized image $I_{t'_{0}\rightarrow t}$ with respect to the target image.
Now, we bilinearly sample from the synthesized image as \begin{equation}
    I_{(t'_{0}\rightarrow t)\rightarrow t} = I_{t'_{0}\rightarrow t}\langle proj({D}_t, {T}^{res\;-1}_{(t'_0\rightarrow  t)\rightarrow t}, K) \rangle.
\label{eq:res_warp}
\end{equation}
Once we obtain a new synthesized view, we can continue to estimate the next residual poses for next view synthesis. 
For simplicity of notation in Equation~\eqref{eq:res_warp}, we replace the subscript $t'_{0}\rightarrow t$ with $t'_{1}$ to indicate that one warping transformation is applied, and similarly for the $i^{\rm th}$ transformation.
Thus, a general form of Equation~\eqref{eq:res_warp} is defined by
\begin{equation}
    I_{t'_{i}\rightarrow t} = I_{t'_{i}}\langle proj({D}_t, {T}^{res-1}_{t'_i\rightarrow t}, K) \rangle, i=0,1,\cdots\;.
\label{eq:res_warp_formal}
\end{equation}
After we estimate multiple residual poses, the camera pose of source image $I_t'$ with respect to the target image $I_t$ can be written as ${T}_{t\rightarrow t'}={T}^{-1}_{t'\rightarrow t}$ where
\begin{equation}
    {T}_{t'\rightarrow t} = \prod_{i}{T}_{t'_i\rightarrow t}, i=\cdots,k,\cdots,1,0 \;.
\label{eq:pose}
\end{equation}
By iteratively estimating residual poses, we expect to obtain more accurate camera poses compared with the pose predicted from a single-stage pose network, 
so that a more accurate photometric reprojection loss can be built up for better depth prediction.

\section{Experiments}

\noindent\textbf{Datasets.} We evaluate the proposed framework \textbf{\textit{MonoIndoor}} on two challenging indoor datasets: the EuRoC MAV~\cite{schonberger2016structure} dataset, the NYUv2 depth dataset~\cite{Silberman:ECCV12} and RGB-D 7-Scenes dataset~\cite{Shotton_2013_CVPR}. 

\noindent\textbf{Evaluation Metrics.} For evaluation, we follow~\cite{eigen2014depth} to use the mean absolute relative error (AbsRel), root mean squared error (RMS), and the accuracy under threshold ($\delta_{i}<1.25^{i}, i=1,2,3$) on both datasets.

\noindent\textbf{Implementation Details.} We implement our model using PyTorch~\cite{NEURIPS2019_9015}. In the depth factorization module, we use the same depth network as in~\cite{godard2019digging}; for the scale network, we use two basic residual blocks followed by three fully-connected layers with a dropout layer in-between. The dropout rate is set to 0.5. In the residual pose module, we let the residual pose networks use a common architecture~\cite{godard2019digging} which consists of a shared pose encoder and an independent pose regressor.  Each experiment is trained for 40 epochs using the Adam~\cite{kingma2015adam} optimizer and the learning rate is set to $10^{-4}$ for the first 20 epochs and it drops to $10^{-5}$ for remaining epochs. The smoothness term $\tau$ and consistency term $\gamma$ are set as 0.001 and 0.05, respectively.

\subsection{EuRoC MAV Dataset}
The EuRoC MAV Dataset~\cite{schonberger2016structure} contains 11 video sequences captured in two main scenes, a machine hall and a vicon room. Sequences are categorized as \textit{easy}, \textit{medium} and \textit{difficult} according to the varying illumination and camera motions.
For the training, we use three sequences of ``Machine hall" (MH\_01, MH\_02, MH\_04) and two sequences of ``Vicon room'' (V1\_01 and V1\_02). Images are rectified with provided camera intrinsics to remove image distortion. During training, images are resized to 512$\times$256. Following~\cite{gordon2019depth}, we use the Vicon room sequence V2\_01 for testing where the ground-truth depths are generated by projecting Vicon 3D scans onto the image planes.

\subsubsection{Ablation Study}
\label{sec:euroc_ablation}
We perform ablation studies for our design choices of the depth factorization module on the EuRoC MAV dataset. Firstly, we consider the following designs as the backbone of our scale network: I) a pre-trained ResNet-18~\cite{He_2016_CVPR} followed by a group of Conv-BN-ReLU layers; II) a pre-trained ResNet-18~\cite{He_2016_CVPR} followed by two residual blocks; III) a lightweight network with two residual blocks which shares the feature maps from the depth encoder as input. These three choices are referred to as the ScaleCNN, ScaleNet and ScaleRegressor, respectively in Table~\ref{tab:eurco_quan_ablation}. Next, we validate the effectiveness of adding new components into our backbone design. As described in Section~\ref{sec:depth_scale_factorization}, we mainly integrate two sub-modules: i) a self-attention block and ii) a probabilistic scale regression block.

\begin{table}[!t]
    \caption{Ablation results of design choices and the effectiveness of components in the depth factorization module of our model (\textbf{MonoIndoor})  on EuRoC~\cite{schonberger2016structure}. Porb. Reg.: the probabilistic scale regression block. Note: here we also use the residual pose estimation module when experimenting with different network designs for the depth factorization module.}
    \label{tab:eurco_quan_ablation}
    \centering
    \resizebox{0.48\textwidth}{!}{
    \begin{tabular}{c|c|c|c|c|c|c|c}
    \hline
    \multirow{2}{*}{Network Design} & 
    \multirow{2}{*}{Attention} &
    \multirow{2}{*}{\tabincell{c}{Prob.\\Reg.}} &
    \multicolumn{2}{c|}{Error Metric} &  \multicolumn{3}{c}{Accuracy Metric} \\
    \cline{4-8}
    ~ & ~ & ~ & AbsRel & RMSE  & $\delta_1$ & $\delta_2$ & $\delta_3$ \\
    \hline
    I. ScaleCNN & \cmark & \cmark & 0.140 & 0.518 & 0.821 & 0.956 & 0.985 \\
    II. ScaleNet & \cmark & \cmark & 0.141 & 0.519 & 0.817 & 0.959 & 0.988 \\
    \hline
    III. ScaleRegressor & \xmark & \xmark & 0.139 & 0.508 & 0.817 & 0.960 & 0.987 \\
    III. ScaleRegressor & \cmark & \xmark & 0.135 & 0.501 & 0.825 & 0.964 & 0.989 \\
    III. ScaleRegressor & \cmark & \cmark & 0.125 & 0.466 & 0.840 & 0.965 & 0.993 \\
    \hline
    \end{tabular}
    }
\end{table}

As shown in Table~\ref{tab:eurco_quan_ablation}, the best performance is achieved by ScaleRegressor that uses self-attention and probabilistic scale regression. It proves that sharing features with the depth encoder is beneficial to scale estimation.
Comparing the results of three ScaleRegressor variants, the performance gradually improves as we add more components (\ie., attention and Prob. Reg.). Specifically, adding the self-attention block improves the overall performance over the baseline backbone; adding the probabilistic regression block leads to a further improvement, which validates the effectiveness of our proposed sub-modules.

\subsubsection{Quantitative Results}
Since there are not many public results reported on the EuRoC MAV~\cite{schonberger2016structure} dataset, we mainly compare our model with the baseline model Monodepth2~\cite{godard2019digging} and validate the effectiveness of each module of our MonoIndoor.
As shown in Table~\ref{tab:eurco_quan_full}, adding our depth factorization module reduces the AbsRel from 15.7\% to 14.9\%, and  our residual pose module decreases the AbsRel to 14.1\%, which verifies the usefulness of each module. Our full model achieves the best performance across all evaluation metrics. Specifically, compared to Monodepth2, the AbsRel by our {\bf MonoIndoor} is significantly decreased from 15.7\% to 12.5\% and the $\delta_1$ is improved by around 6\%, from 78.6\% to 84.0\%.

\begin{table}[!h]
    \caption{Ablation results of our MonoIndoor and quantitative comparison with the baseline on the test sequence V2\_01 of EuRoC. Best results are in \textbf{bold}.}
    \label{tab:eurco_quan_full}
    \centering
    \resizebox{0.48\textwidth}{!}{
    \begin{tabular}{c|c|c|c|c|c|c|c}
    \hline
    \multirow{2}{*}{Method} & 
    \multirow{2}{*}{\tabincell{c}{Depth\\Factorization}}  & 
    \multirow{2}{*}{\tabincell{c}{Residual \\Pose}} & 
    \multicolumn{2}{c|}{Error Metric} &  \multicolumn{3}{c}{Accuracy Metric} \\
    \cline{4-8}
    ~ & ~ & ~ & AbsRel & RMSE  & $\delta_1$ & $\delta_2$ & $\delta_3$ \\
    \hline
    Monodepth2~\cite{godard2019digging} & \xmark & \xmark & 0.157  & 0.567  & 0.786 & 0.941 & 0.986 \\
    MonoIndoor & \cmark & \xmark & 0.149 & 0.535  & 0.805 & 0.955 & 0.987 \\
    MonoIndoor & \xmark & \cmark & 0.141 & 0.518 & 0.815 & 0.961 & 0.991 \\
    MonoIndoor & \cmark & \cmark & \textbf{0.125} & \textbf{0.466} & \textbf{0.840} & \textbf{0.965} & \textbf{0.993} \\
    \hline
    \end{tabular}
    }
\end{table}

\begin{figure}[!t]
\begin{center}
\includegraphics[width=\linewidth]{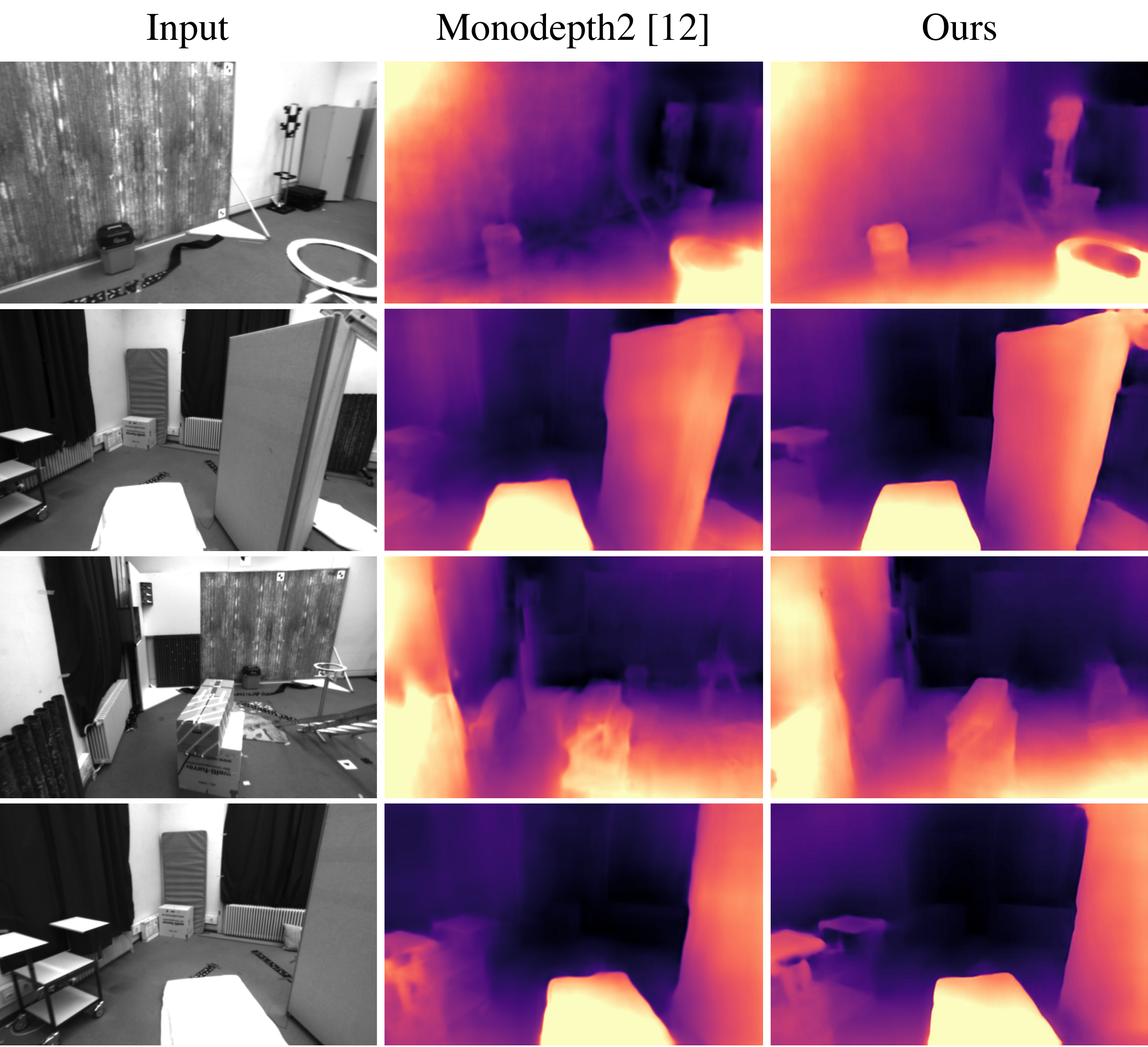}
\end{center}
\caption{Qualitative comparison of depth prediction on EuRoC. Our model produces more accurate and cleaner depth maps.}
\label{fig:euroc_qua_full}
\end{figure}

\subsubsection{Qualitative Results}
Figure~\ref{fig:euroc_qua_full} gives a qualitative comparison of depth maps predicted by Monodepth2~\cite{godard2019digging} and our \textbf{MonoIndoor}. 
From Figure~\ref{fig:euroc_qua_full}, it is clear that the depth maps generated by our model are much better than the ones by Monodepth2. For instance, in the first row, our model can predict precise depths for the \textbf{\textit{hole}} region at the right-bottom corner whereas such a hole structure in the depth map by Monodepth2 is missing. Besides, in the second row, our model can predict much sharper depth map of the \textit{\textbf{ladder}} at the right-top area while Monodepth2 cannot. These observations are also consistent with the better quantitative results in Table~\ref{tab:eurco_quan_full}, proving the superiority of our model.

\subsection{NYUv2 Depth Dataset}
In this section, we evaluate our \textbf{MonoIndoor} on the NYUv2 depth dataset~\cite{Silberman:ECCV12} which contains 464 indoor video sequences captured by a hand-held Microsoft Kinect RGB-D camera with a resolution of 640$\times$ 480. We use the official training and validation splits which include 302 and 33 sequences respectively. We rectify the images with provided camera parameters to remove distortions. Following~\cite{zhao2020towards, bian2020unsupervised}, the raw dataset is firstly downsampled 10 times along the temporal dimension to remove redundant frames, resulting in $\sim20K$ images for training. During training, images are resized to 320$\times$256. We use officially provided 654 images with dense labelled depth maps for testing.

\begin{table}[h!]
    \caption{Ablation results of the effectiveness of each module of our MonoIndoor on NYUv2. ``No. Residual Pose Block'' means the number of residual poses we estimate in the residual pose estimation module.}
    \label{tab:nyuv2_quan_ablation}
    \centering
    \resizebox{0.48\textwidth}{!}{
    \begin{tabular}{c|c|c|c|c|c|c|c}
    \hline
    \multirow{2}{*}{Model} &
    \multirow{2}{*}{\tabincell{c}{Depth\\Factorization}}  & 
    \multirow{2}{*}{\tabincell{c}{No. Residual\\Pose Block}} & 
    \multicolumn{2}{c|}{Error Metric} &  \multicolumn{3}{c}{Accuracy Metric} \\
    \cline{4-8}
    ~ & ~ & ~ & AbsRel & RMS & $\delta_1$ & $\delta_2$ & $\delta_3$ \\
    \hline
    Monodepth2~\cite{godard2019digging} & \xmark & 0 & 0.16 & 0.601 & 0.767 & 0.949 & 0.988 \\
    MonoIndoor & \cmark & 0 & 0.152 & 0.576 & 0.792 & 0.951 & 0.987 \\
    \hline
    MonoIndoor & \xmark & 1 & 0.142 & 0.553 & 0.813 & 0.958 & 0.988 \\
    MonoIndoor & \cmark & 1 & \textbf{0.134} & \textbf{0.526} & \textbf{0.823} & \textbf{0.958} & \textbf{0.989} \\
    \hline
    MonoIndoor & \xmark & 2 & 0.141 & 0.548 & 0.814 & 0.958 & 0.988 \\
    MonoIndoor & \cmark & 2 & 0.141 & 0.546 & 0.818 & 0.958 & 0.989 \\
    \hline
    \end{tabular}
    }
\end{table}

\subsubsection{Ablation Study}
\begin{figure*}[h]
\begin{center}
\includegraphics[width=\linewidth]{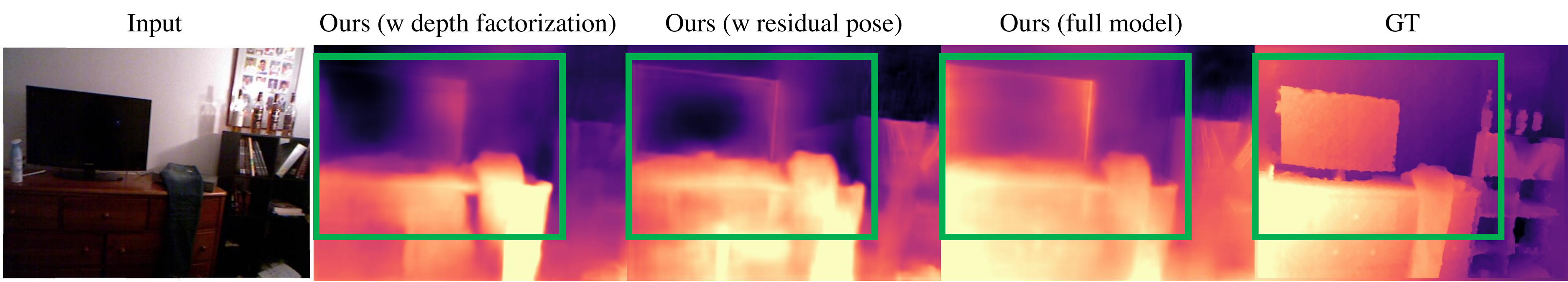}
\end{center}
\caption{Qualitative ablation comparisons of depth prediction on NYUv2. Our full model with both depth factorization and residual pose modules produce better depth maps.}
\label{fig:nyuv2_qua_ablations}
\end{figure*}

We perform another ablation study for the depth factorization module on NYUv2~\cite{Silberman:ECCV12}. In Table~\ref{tab:nyuv2_quan_ablation}, comparing with Monodepth2 which predicts depth without any guidance of global scales, using the depth factorization module with a separate scale network can improve the performance, decreasing the AbsRel from 16\% to 15.2\% and increasing $\delta_1$ to 79.2\%. Next, we experiment to validate the effectiveness of the residual pose estimation module. Comparing the rows in Table~\ref{tab:nyuv2_quan_ablation}, by adding the residual pose estimation module with one residual pose block, we observe an improved performance from 16.0\% down to 14.2\% for the AbsRel and from 76.7\% up to 81.3\% for $\delta_1$. Furthermore, by applying both the depth factorization module and the residual pose estimation module (\ie, our full \textbf{MonoIndoor}), significant improvements can be achieved across all evaluation metrics. For instance, the AbsRel is reduced to 13.4\% and the $\delta_1$ is increased to 82.3\%. These ablation results clearly prove the effectiveness of the proposed modules, which also align with the qualitative results in Figure~\ref{fig:nyuv2_qua_ablations} where we visualize predictions on NYUv2 by our proposed modules. However, referring to the last two rows, when adding more residual pose blocks and training with/without the depth factorization module, the performance does not significantly improve or even becomes worse. We will leave the investigation of this phenomenon for future work.

\begin{figure*}[h]
\begin{center}
\includegraphics[width=\linewidth]{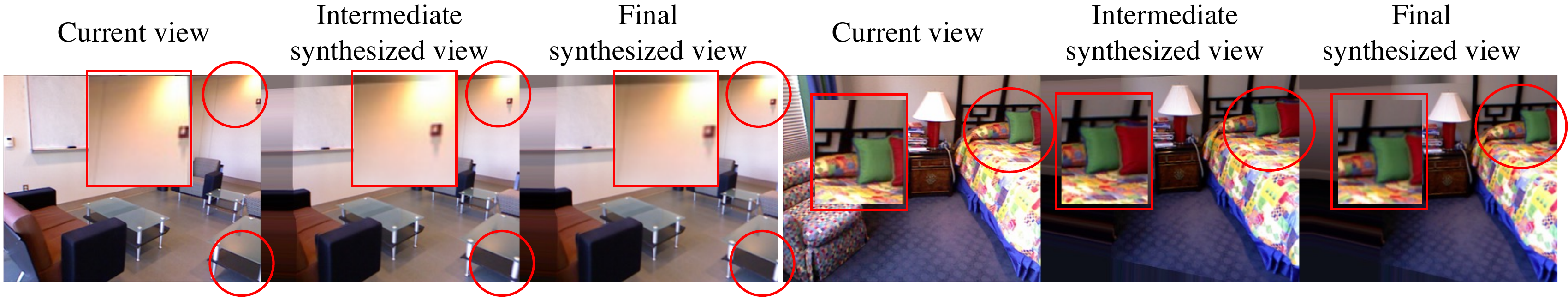}
\end{center}
\caption{Intermediate synthesized views on NYUv2.}
\label{fig:nyuv2_intermediate_views}
\end{figure*}

We further visualize intermediate and final synthesized views compared with the current view on NYUv2 in the Figure~\ref{fig:nyuv2_intermediate_views}. Highlighted regions show that final synthesized views are better than the intermediate synthesized views and closer to the current view.

\begin{table}[!t]
    \caption{Comparison of our method to existing supervised and self-supervised methods on NYUv2~\cite{Silberman:ECCV12}. Best results among supervised and self-supervised methods are in \textbf{bold}. }
    \label{tab:nyuv2_quan_full}
    \centering
    \resizebox{0.48\textwidth}{!}{
    \begin{tabular}{c|c|c|c|c|c|c}
    \hline
    \multirow{2}{*}{Methods} & 
    \multirow{2}{*}{Supervision} & \multicolumn{2}{c|}{Error Metric} & \multicolumn{3}{c}{Accuracy Metric} \\
    \cline{3-7}
    ~ & ~ & AbsRel & RMS & $\delta_1$ & $\delta_2$ & $\delta_3$ \\
    \hline
    Make3D~\cite{saxena2008make3d} & \cmark & 0.349 & 1.214 & 0.447 & 0.745 & 0.897 \\
    Depth Transfer~\cite{Karsch:TPAMI:14} & \cmark & 0.349 & 1.210 & - & - & - \\
    Liu~\etal~\cite{Liu_2014_CVPR} & \cmark & 0.335 & 1.060 & - & - & - \\
    Ladicky~\etal~\cite{Ladicky_2014_CVPR} & \cmark & - & - & 0.542 & 0.829 & 0.941 \\
    Li~\etal~\cite{Li_2015_CVPR} & \cmark & 0.232 & 0.821 & 0.621 & 0.886 & 0.968 \\
    Roy~\etal~\cite{Roy_2016_CVPR} & \cmark & 0.187 & 0.744 & - & - \\
    Liu~\etal~\cite{Liu_2015_CVPR} & \cmark & 0.213 & 0.759 & 0.650 & 0.906 & 0.976 \\
    Wang~\etal~\cite{Wang_2015_CVPR} & \cmark & 0.220 & 0.745 & 0.605 & 0.890 & 0.970 \\
    Eigen~\etal~\cite{Eigen_2015_ICCV} & \cmark & 0.158 & 0.641 & 0.769 & 0.950 & 0.988 \\
    Chakrabarti~\etal~\cite{NIPS2016_f3bd5ad5} & \cmark & 0.149 & 0.620 & 0.806 & 0.958 & 0.987 \\
    Laina~\etal~\cite{laina2016deeper} & \cmark & 0.127 & 0.573 & 0.811 & 0.953 & 0.988 \\
    Li~\etal~\cite{Li_2017_ICCV} & \cmark & 0.143 & 0.635 & 0.788 & 0.958 & 0.991 \\
    DORN~\cite{Fu_2018_CVPR} & \cmark & 0.115 & 0.509 & 0.828 & 0.965 & 0.992 \\
    VNL~\cite{Yin_2019_ICCV} & \cmark & 0.108 & 0.416 & \textbf{0.875} & \textbf{0.976} & \textbf{0.994} \\
    Fang~\etal~\cite{Fang_2020_WACV} & \cmark & \textbf{0.101} & \textbf{0.412} & 0.868 & 0.958 & 0.986 \\
    \hline
    \hline
    Zhou~\etal~\cite{zhou2019moving} & \xmark & 0.208 & 0.712 & 0.674 & 0.900 & 0.968 \\
    Zhao~\etal~\cite{zhao2020towards} & \xmark & 0.189 & 0.686 & 0.701 & 0.912 & 0.978 \\
    Monodepth2~\cite{godard2019digging} & \xmark & 0.160 & 0.601 & 0.767 & 0.949 & 0.988 \\
    Bian~\etal~\cite{bian2020unsupervised} & \xmark & 0.147 & 0.536 & 0.804 & 0.950 & 0.986 \\
    \hline
    \textbf{MonoIndoor}(Ours) & \xmark & \textbf{0.134} & \textbf{0.526} & \textbf{0.823} & \textbf{0.958} & \textbf{0.989} \\
    \hline
    \end{tabular}
    }
\end{table}

\begin{table*}[!t]
    \caption{Comparison of our method to latest self-supervised methods on RGB-D 7-Scenes~\cite{Shotton_2013_CVPR}. Best results are in \textbf{bold}}.
    \label{tab:7scenes_quan_full}
    \centering
    \resizebox{0.7\textwidth}{!}{
    \begin{tabular}{c|c|c|c|c|c|c|c|c}
    \hline
    \multirow{3}{*}{Scenes} &
    \multicolumn{4}{c|}{Bian~\etal~\cite{bian2020unsupervised}} &
    \multicolumn{4}{c}{\textbf{MonoIndoor} (Ours)} \\
    \cline{2-9}
    ~ & \multicolumn{2}{c|}{Before Fine-tuning} &
    \multicolumn{2}{c|}{After Fine-tuning} &
    \multicolumn{2}{c|}{Before Fine-tuning} &
    \multicolumn{2}{c}{After Fine-tuning} \\
    \cline{2-9}
    ~ & AbsRel & Acc $\delta_1$ & AbsRel & Acc $\delta_1$ & AbsRel & Acc $\delta_1$ & AbsRel & Acc $\delta_1$ \\
    \hline
    Chess & 0.169 & 0.719 & 0.103 & 0.880 & 0.157 & 0.750 & \textbf{0.097} & \textbf{0.888} \\
    Fire & 0.158 & 0.758  & 0.089 & 0.916 & 0.150 & 0.768 & \textbf{0.077} & \textbf{0.939} \\
    Heads & 0.162 & 0.749 & 0.124 & 0.862 & 0.171 & 0.727 & \textbf{0.106} & \textbf{0.889} \\
    Office & 0.132 & 0.833 & 0.096 & 0.912 & 0.130 & 0.837 & \textbf{0.083} & \textbf{0.934} \\
    Pumpkin & 0.117 & 0.857 & 0.083 & \textbf{0.946} & 0.102 & 0.895 & \textbf{0.078} & 0.945 \\
    RedKitchen & 0.151 & 0.78 & 0.101 & 0.896 & 0.144 & 0.795 & \textbf{0.094} & \textbf{0.915} \\
    Stairs & 0.162 & 0.765 & 0.106 & 0.855 & 0.155 & 0.753 & \textbf{0.104} & \textbf{0.857} \\
    \hline
    \end{tabular}
    }
\end{table*}

\subsubsection{Quantitative Results}

We present the quantitative results of our model \textbf{MonoIndoor} and both state-of-the-art (SOTA) supervised and self-supervised methods on NYUv2 in Table~\ref{tab:nyuv2_quan_full}. It shows that our model outperforms previous self-supervised SOTA methods, reaching the best results across all metrics. Specifically, compared to a recent self-supervised method by Bian~\etal~\cite{bian2020unsupervised} which removes rotations via ``weak rectification'', our method reduces AbsRel by 1.3\% and increases $\delta_1$ by 1.9\%, reaching an AbsRel of 13.4\%  and $\delta_1$ of 82.3\%.  In addition to that, our model outperforms a group of supervised methods and close the performance gap between the self-supervised methods and fully-supervised methods. 

\subsubsection{Qualitative Results}

\begin{figure*}[!t]
\begin{center}
\includegraphics[width=1.0\linewidth]{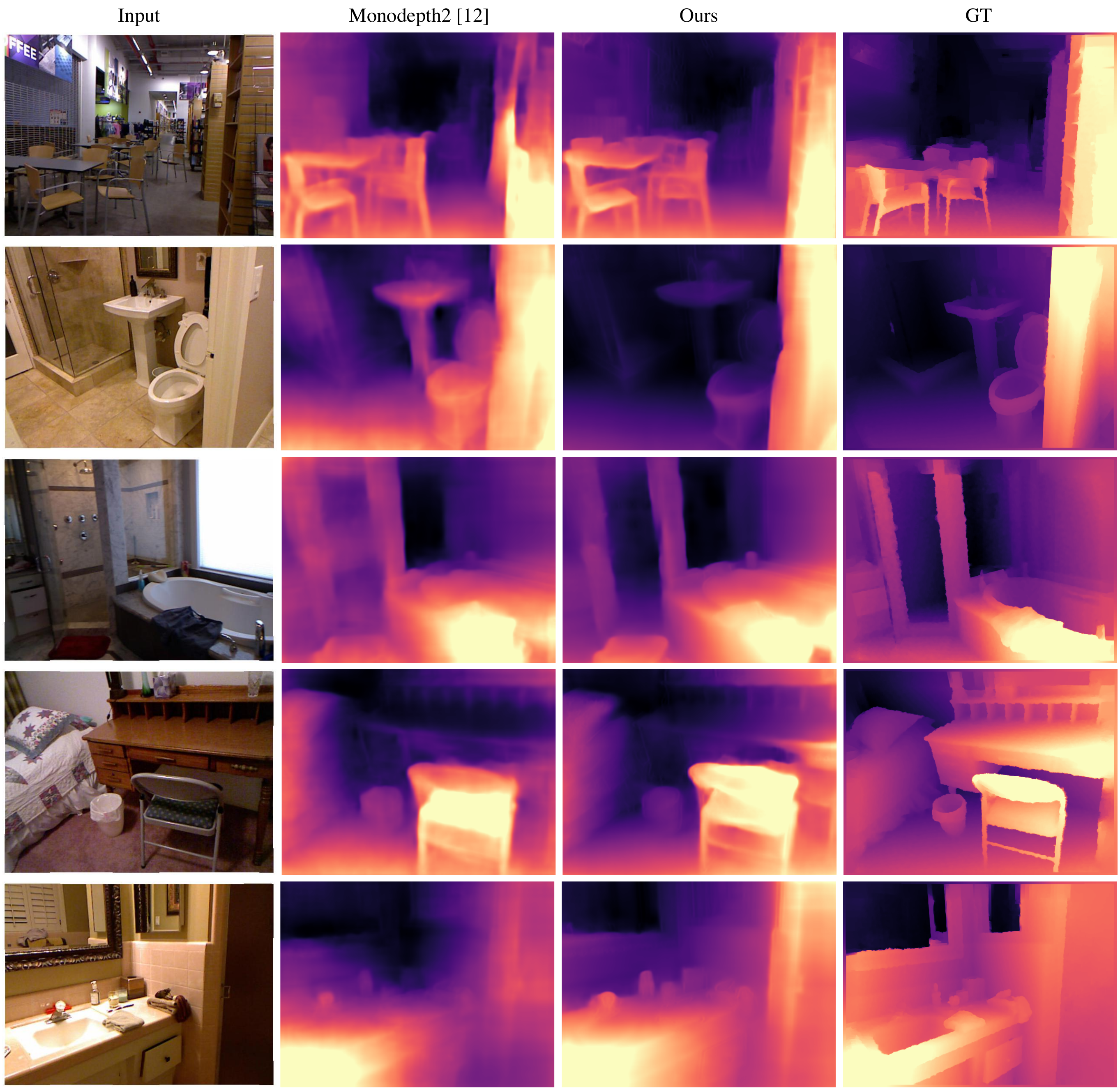}
\end{center}
\caption{Qualitative comparison on NYUv2~\cite{Silberman:ECCV12}. Compared with Monodepth2~\cite{godard2019digging}, our model produces accurate depth maps (in the third column) that are closer to the ground-truth.}
\vspace{-0.1cm}
\label{fig:nyuv2_qua_full}
\end{figure*}

Figure~\ref{fig:nyuv2_qua_full} visualizes the predicted depth maps on NYUv2. Compared with the results from the Monodepth2~\cite{godard2019digging}, depth maps predicted from our model (\textbf{MonoIndoor}) are more precise and closer to the ground-truth. For instance, looking at the third column in the first row, the depth in the region of \textbf{\textit{chairs}} predicted from our model is much sharper and cleaner, 
being close to the ground truth (the last column). On the rightmost area of the same image where there is a \textbf{\textit{shelf}}, our model can produce better depth predictions that reflect its shape. These observations are consistent with our quantitative results in Table~\ref{tab:nyuv2_quan_full}.

\subsection{RGB-D 7-Scenes Dataset}
In this section, we evaluate our \textbf{MonoIndoor} on the RGB-D 7-Scenes dataset \cite{Shotton_2013_CVPR} which contains several video sequences with 500-1000 frame in each sequence. All scenes are recorded using a handheld Kinect RGB-D camera at 640×480 resolution. We use the official train/test split. Following~\cite{bian2020unsupervised}, for training, we first pre-train our \textbf{MonoIndoor} on NYUv2 dataset, and then fine-tune the model on this dataset; for testing, we extract one image from every 30 frames. Images are resized to $320\times256$ during training.

\subsubsection{Quantitative Results}

We present the quantitative results of our model \textbf{MonoIndoor} and latest state-of-the-art (SOTA) self-supervised methods on 7-Scenes in Table 5. It shows that our model outperforms~\cite{bian2020unsupervised} on most scenes before and after fine-tuning, demonstrating better generalizability and capability of our model. Specifically, compared to a recent self-supervised method by Bian et al.~\cite{bian2020unsupervised}, on the scene ``Fire", our method reduces AbsRel by 1.2\% and increases $\delta_1$ by 2.3\%, reaching an AbsRel of 7.7\% and $\delta_1$ of 93.9\%; on the scene "Heads",our method reduces AbsRel by 1.8\% and increases $\delta_1$ by 2.7\%, reaching an AbsRel of 10.6\% and $\delta_1$ of 88.9\%.

\section{Conclusions}
In this work, we have presented a novel monocular self-supervised depth estimation model, namely \textbf{MonoIndoor}, to study good practices towards predicting accurate depth maps in indoor environments. We first introduce the \textit{depth factorization module} to jointly learn a global scale factor and a relative depth map from an input image. To estimate accurate camera poses for novel view synthesis, we propose a \textit{residual pose estimation module} that decomposes a global pose into an initial pose and one or a few residual poses, which in turn improves the depth model. We have shown that our model achieves the state-of-the-art performance among the self-supervised methods on two challenging indoor datasets, \ie, EuRoC and NYUv2. 

It is to be noted that our depth factorization module is in itself agnostic to the types of supervision, so it may also be helpful for supervised depth prediction. In the future, we plan to investigate its effectiveness in a supervised setup. Another interesting future direction would be to train our method on multiple datasets with various depth ranges and then test it for zero-shot cross-dataset transfer as in~\cite{ranftl2019towards}.

\section*{Appendix}
\subsection*{Network Details}
In the depth factorization module, we use the same depth network of an auto-encoder structure as in~\cite{godard2019digging} to predict the relative depth, and employ a scale network consisting of an encoder and a regressor. The encoder of the scale network is shared with the depth encoder and the architecture of the scale regressor is described in Table~\ref{tab:scale_regressor}. In the residual pose module, we use one pose network and one residual pose network, both of which share the same structure. The residual pose network shares parameters in the encoder with the pose network but learns independent parameters in its pose prediction head.

\begin{table}[h]
    \caption{Scale regressor architecture. Here \textbf{chns} is the number of ouput channels, \textbf{k} is the kernal size, \textbf{s} is the stride, \textbf{res} is the downscaling factor for each layer with respect to the input image, and \textbf{input} is the input to each layer.}
    \label{tab:scale_regressor}
    \centering
    \resizebox{0.47\textwidth}{!}{
    \begin{tabular}{|c|c|c|c|c|}
        \hline
        \multicolumn{5}{|c|}{\textbf{Scale Regressor}} \\
        \hline
        \textbf{Block} & \textbf{layer} & \textbf{chns-k-s} & \textbf{res} & \textbf{input}  \\
        \hline
        Attention & \tabincell{c}{(query)\\(key)\\(value)\\} & \tabincell{c}{(512, 1, 1)\\(512, 1, 1)\\(512, 1, 1)\\} & 32 & \tabincell{c}{(econv5)\\(econv5)\\(econv5)\\} \\
        \hline
        ConvBlock1 & \tabincell{c}{(convs1\_0)\\(convs1\_1)} & \tabincell{c}{(512, 3, 1)\\(512, 3, 1)} & 32 & Attention \\
        \hline
        ConvBlock2 & convs2\_0 & 1024, 1, 2 & 64 & ConvBlock1 \\
        \hline
        ConvBlock3 & \tabincell{c}{(Convs3\_0)\\(Convs3\_1)} & \tabincell{c}{(1024, 3, 1)\\(1024, 3, 1)} & 64 & ConvBlock2 \\
        \hline
        \multicolumn{5}{|c|}{FC1-1024-Dropout} \\
        \hline
        \multicolumn{5}{|c|}{FC2-1024-Dropout} \\
        \hline
        \multicolumn{5}{|c|}{Scale Regression} \\
        \hline
    \end{tabular}
    }
\end{table}

\begin{figure}[!t]
\begin{center}
\includegraphics[width=0.98\linewidth]{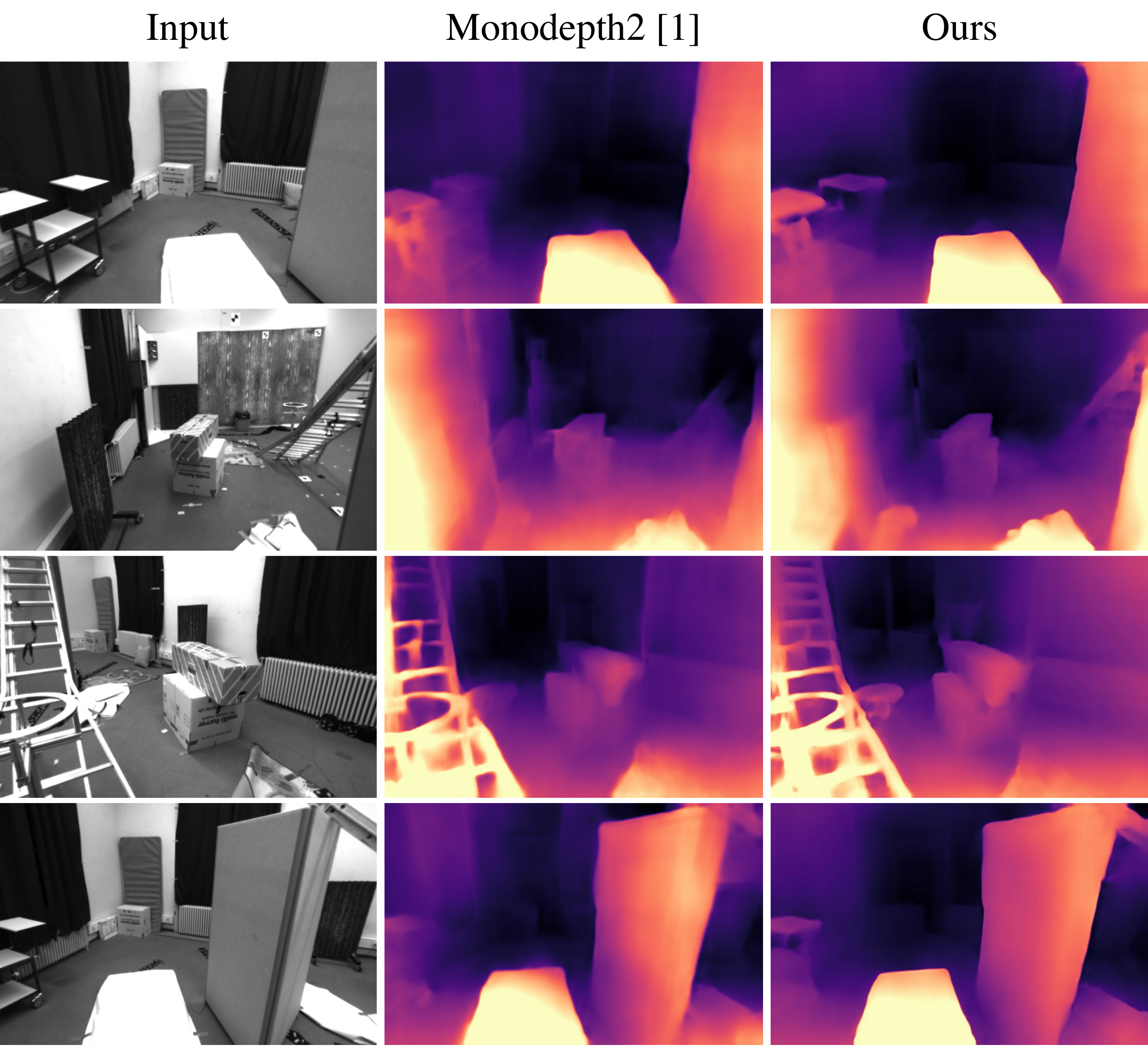}
\end{center}
\caption{Additional qualitative comparison on EuRoC MAV\cite{schonberger2016structure}.}
\label{fig:euroc_supp_qua_full}
\end{figure}

\subsection*{Odometry Evaluation}
In Table~\ref{tab:euroc_pose_eval}, we evaluate the proposed residual pose estimation module on the test sequences V1\_03 and V2\_01 of the EuRoC MAV~\cite{schonberger2016structure}. We follow~\cite{zhan2019dfvo} to evaluate relative camera poses estimated by our residual pose estimation module. We use the following evaluation metrics: absolute trajectory error (ATE) which measures the root-mean square error between predicted camera poses and ground-truth, and relative pose error (RPE) which measures frame-to-frame relative pose error in meters and degrees, respectively. As shown in Table~\ref{tab:euroc_pose_eval}, on both two test sequences, compared with the baseline model Monodepth2~\cite{godard2019digging} which employs one-stage pose network, using our residual pose estimation module leads to improved relative pose estimation across all evaluation metrics. Specifically, on the sequence V1\_03, the ATE by our \textbf{MonoIndoor} is significantly decreased from 0.0681 meters to 0.052 meters and PRE(\textdegree) is reduced by around half, from 1.3237\textdegree  to 0.7179\textdegree.

\begin{table}[!t]
    \caption{Odometry results on the EuRoC MAV~\cite{schonberger2016structure} test set. Results show the average absolute trajectory error(ATE), and the relative pose error(RPE) in meters and degrees, respectively. Seq.: sequence name. }
    \label{tab:euroc_pose_eval}
    \centering
    \resizebox{0.47\textwidth}{!}{
    \begin{tabular}{|c|c|c|c|c|}
    \hline
    Seq. & Methods &
    ATE(m) & RPE(m) & RPE(\textdegree) \\
    \hline
    \multirow{2}{*}{V1\_03} & Monodepth2~\cite{godard2019digging} & 0.0681 & 0.0686 & 1.3237 \\
    \cline{2-5}
    ~ & \textbf{MonoIndoor}(Ours) & \textbf{0.052} & \textbf{0.0637} & \textbf{0.7179} \\
    \hline
    \hline
    \multirow{2}{*}{V2\_01} & Monodepth2~\cite{godard2019digging} & 0.0266 & 0.0199 & 1.1985 \\
    \cline{2-5}
    ~ & \textbf{MonoIndoor}(Ours) & \textbf{0.0222} & \textbf{0.0109} & \textbf{1.1974} \\
    \hline
    \end{tabular}
    }
\end{table}

\begin{figure*}[!t]
\begin{center}

\includegraphics[width=0.97\linewidth]{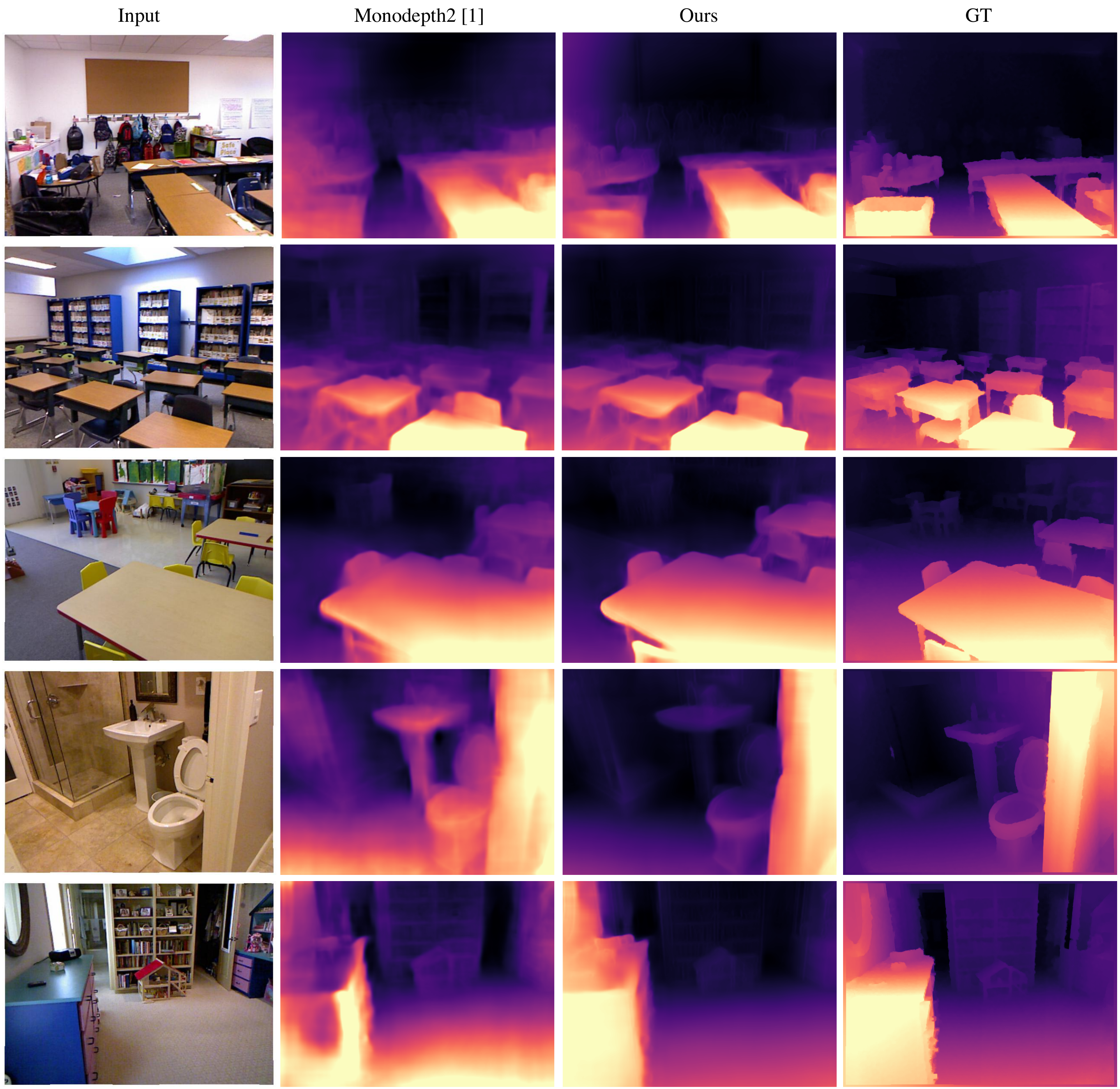}
\end{center}
\caption{Additional qualitative comparison on NYUv2~\cite{Silberman:ECCV12}.}
\vspace{-0.2cm}
\label{fig:nyuv2_supp_qua_full}
\end{figure*}

\subsection*{Additional Qualitative Results}
We include additional qualitative results on both the EuRoC and NYUv2 test sets in Figure~\ref{fig:euroc_supp_qua_full} and Figure~\ref{fig:nyuv2_supp_qua_full}, respectively. From both figures, we can see that our models generate depth maps of higher quality.

{\small
\bibliographystyle{ieee_fullname}
\bibliography{egbib}
}

\end{document}